\RequirePackage[no-math]{fontspec}
\defaultfontfeatures{Ligatures=TeX}
\documentclass{clv2025}
\usepackage{setspace}
\usepackage{mathpazo}  
\usepackage{microtype}
\usepackage{multirow}
\usepackage{booktabs}
\usepackage[table]{xcolor}
\usepackage{amsmath}
\usepackage[caption=false]{subfig}
\usepackage{adjustbox}
\usepackage[multiple]{footmisc}
\usepackage{enumitem}
\usepackage{xurl}
\usepackage{graphicx}
\usepackage{xcolor}
\usepackage[T1]{fontenc}

\definecolor{light-gray}{gray}{0.95} 

\definecolor{shadecolor}{gray}{0.95}
\definecolor{pivot}{HTML}{9C2A20}
\definecolor{gl}{gray}{0.45}

\newcommand{\p}[1]{\textcolor{pivot}{#1}}
\newcommand{\g}[1]{\textcolor{gl}{#1}}

\begin{document}

\title{Last Words: Boring Problems Revisited}
\author{Huayang Li\thanks{
8916-5 Takayama-cho, Ikoma, Nara 630-0192, Japan. E-mail: {\tt
  li.huayang.lh6@is.naist.jp}},
Richard Sproat\thanks{
Azabudai Hills Mori JP Tower 22F, 1-3-1 Azabudai, Minato-ku, Tokyo 106-0041,
Japan. E-mail: {\tt rws@sakana.ai}}
\affil{Nara Institute of Science and Technology and Sakana.ai}
}

\issue{1}{1}{2026}

\dochead{}

\runningtitle{Last Words: Boring Problems Revisited}
\runningauthor{Li \& Sproat}

\maketitle

\begin{abstract}
In 2022, the second author published a \emph{Last Words} piece in
\emph{Computational Linguistics} in which he
pointed out that, despite the advances in AI up to that point, there were still
some problems in natural language processing that on the one hand seemed trivial
and perhaps boring, but on the other were hard to solve using purely neural
methods. One such problem discussed in that context was \emph{text
normalization}, and a particular subproblem of that, namely \emph{number name
reading}. The piece came out just a few months before the first release of
ChatGPT, and needless to say, a lot has happened since then. Indeed, as we
document here, the problem (of number name reading) has practically been
\emph{solved}, at least by one of the latest frontier models. However, we
believe that there are still open issues for such limited domain, constrained
and limited resource problems, and that in some future machine learning
approach, the massive resources of the latest frontier models would not be
needed to solve them.

This paper was to be a follow-up to the 2022 piece. In those days, under the
previous editor, the journal followed a time-honored tradition of allowing
authors the opportunity to resubmit papers, unless the reviews were uniformly
bad. In the case of this paper, we received two reviews, one which was ``revise
and resubmit'' and the other ``minor revisions''.  We note that it would have
been perfectly possible to address the concerns of the less positive reviewer.
But the editors decided to summarily reject it anyway, with no option to
resubmit. Times have changed. But we do thank the two reviewers who wasted their
time providing thoughtful reviews.
\end{abstract}

Four years ago one of the authors published a \emph{Last Words} piece in which
he offered the following challenge \citep{Sproat:22}:
\begin{quote}
I issue a challenge: demonstrate that a purely neural approach to the problem
of \emph{text normalization} is possible. Various groups have tried, but so far
nobody has eliminated the problem of \emph{unrecoverable errors}, errors where,
due to insufficient training data or faulty generalization, the system
substitutes some other reading for the correct one.
\end{quote}
The context for this challenge, published soon before the first release of
ChatGPT, was the author's experience in one of the less `exciting', but
nonetheless important problems of natural language processing:
\begin{quote}
Though text normalization is hardly an `exciting' problem, I argue that until
one can solve `boring' problems like that using purely AI methods, one cannot
claim that AI is a success.
\end{quote}
Text normalization, the conversion of expressions such as numbers, dates, times,
abbreviations and so forth into some normalized form---e.g. the words that would
actually be spoken in the case of text-to-speech synthesis, continued to be a
challenge even into the neural age. Previous work that the author had been
involved in, such as \citet{Zhang:EtAl:19}, had failed to solve the problem using
purely neural methods and in particular failed to produce systems that were on a
par with carefully hand-crafted systems such as Kestrel \citep{Ebden:Sproat:14}.

Perhaps because the problem \emph{was} so boring, it received very little
attention. One response was \citet{Zhang:EtAl:24}, published after the release
of ChatGPT, which showed that significant gains could be made, but in that work
it took the then latest version of GPT (4.0) to achieve results on a par with
Kestrel on the Kaggle Text Normalization Challenge \citep{Sproat:Gorman:18},
and did not eliminate the problem of unrecoverable errors---fluent, but semantically incorrect verbalizations.

In \citet{Sproat:22}, there was a particular discussion of the problem of number
naming, i.e. the problem of converting from digit sequences into how they would
be verbalized as numbers: \emph{123} would be \emph{one hundred twenty three} in
standard American English, and \emph{1,234,567} would be \emph{one million, two
hundred thirty four thousand, five hundred sixty seven}. Subsequent to the
publication of \citet{Sproat:22} and \citet{Zhang:EtAl:24}, the second author
continued to experiment with large language models, such as the latest Gemini,
and found that while for English the problem (mostly) seemed to be
solved,\footnote{We say \emph{mostly}, because it is not possible to test how
the system would verbalize every possible number, and because in the authors'
experience, neural models tend to make mistakes in rather unpredictable patterns
so that, for example, for no obvious reason, a model might read \emph{1,234,567}
as \emph{one million, two hundred thirty four thousand, five hundred \textbf{seventy
six}}.} for lower-resource languages problems remained, and even fine-tuning or
in-context learning with many-shot examples did not completely fix them.

For high resource languages like English, the model would have been exposed to
potentially thousands of examples of number readings, which brings up the
problem of how much data a model \emph{should} need in order to learn to read
numbers. As \citet{Sproat:22} pointed out, humans do not learn to read numbers
by being exposed to thousands of examples. Rather they learn the basic numbers,
learn to count from 1 to 100, are exposed to a few larger numbers and their
construction and given that, and possibly some special targeted training, learn
the number name system of their language.  A good model of the amount of data
that a model \emph{should} need in order to learn number names is the kind of
grammatical information that is given on sites such as Omniglot, which gives
grammatical information for number systems, among other things. The site gives
the basic words along with examples of complex constructions (e.g. how to form
decade plus unit combinations, or multiples of thousand). While this is not
always the case, it is frequently true that this is all the information one
needs in order to be able to read numbers in the language.  So a model that
mimics humans would have enough `smarts' to be able to make use of just that
amount of information, and do a near perfect job of reading numbers that it had
not seen.  Finite-state `learning' models such as those presented in
\citet{Gorman:Sproat:16,Ritchie:EtAl:19} were able to do this since they
incorporated prior knowledge about the domain (though human expertise was
still needed to write various linguistic fix-up rules).  But in the second
author's experience, this capability eluded LLMs.

Until now.

Part of the problem with testing LLMs on this domain is that, even for
low-resource languages, we cannot be sure what data the system will have
seen. To circumvent that issue, we invented six constructed languages that share
a common base vocabulary, but differ in details of how number names are built,
showing a range of complexity in number name formation that is reflective of
what one actually finds in number-name systems.  In all experiments we focus on
number names between 1 and 999,999 and so the discussion will be limited to this
range.  All of the data and code for constructing these languages, the prompts
used for the LLMs, and the results reported here can be found at
\url{https://github.com/SakanaAI/NumberNames}.

These constructed languages are as follows:
\begin{enumerate}
  \item \textbf{Neoferulian}: The simplest, has words for 1--9, 10, 100 and
    1000. All complex number names are constructed by sums of products of these.
  \item \textbf{Ferulian}: The same as Neoferulian, except that there are also
    morphologically unpredictable words for the teens and decades---more or
    less the same situation as English.
  \item \textbf{Continental Ferulian}: Like Ferulian, but has the same
    decade-unit reordering as is found in modern German and Dutch, and earlier
    forms of English (e.g. \emph{four and twenty}).
  \item \textbf{Archiferulian}: Like Ferulian, but with some
    additional sandhi rules between decades and units.
  \item \textbf{Lurefian}: Like Ferulian, but with reordering of multipliers and
    multiplicands in a way reminiscent of Igbo and some other West African
    languages, where \emph{1,000} is \emph{one thousand} but \emph{2,000} is
    \emph{thousand two}. See, e.g.,
    \url{https://www.omniglot.com/language/numbers/igbo.htm}.
  \item \textbf{Archilurefian}: Like Lurefian, but the same decade-unit flip as
    with Continental Ferulian and with additional sandhi rules after the word
    for hundred.
\end{enumerate}
Representative examples of how to read the number \emph{2022} are given below:
\begin{enumerate}
  \item \textbf{Neoferulian}: \emph{kei kero kei fotik sei}.
  \item \textbf{Ferulian}: \emph{kei kero keifti sei}.
  \item \textbf{Continental Ferulian}: \emph{kei kero sei pa keifti}.
  \item \textbf{Archiferulian}: \emph{kei kero keifti ssei}.
  \item \textbf{Lurefian}: \emph{kero kei keifti sei}.
  \item \textbf{Archilurefian}: \emph{kero kei sei pa keifti}.
\end{enumerate}
Full details can be found in the GitHub repository.

For many-shot prompting, we provided training examples of the numbers for each
language in a JSONL file, including: 1--30, 40, 50, 60, 70, 80, 90, 100, 101,
200, 1,000, 1,001, 10,000, 10,001, 20,000, 100,000, 100,001, 200,000 (though see
below). For Archilurefian, 101--110 were also included to show examples of the
additional sandhi rule mentioned above.

We experimented with two versions of the Claude model family, namely Sonnet 4.5
(dated 2025-09-29), and the most recent version of Opus, namely 4.8, released
May 28, 2026. Both were accessed via Claude Code
(\url{https://claude.com/product/claude-code}) using the Claude command-line
interface. We tested two conditions: pure reasoning, where the LLM was
instructed to come up with its own idea about how the number system worked given
the training samples, and then fill out the correct verbalization for a random
subset of 500 unseen cases. And coding, where the model was first instructed to
write code to implement the number name grammar, and then run that code to
produce the output. The 500 randomly chosen held-out examples were the same
numbers in all experiments.  The LLM was not able to access the correct
verbalization and was explicitly instructed not to access the Web.

\begin{figure}[t]
\begin{center}
\begin{tabular}{c}
\textbf{a} \\
\includegraphics[width=0.6\textwidth]{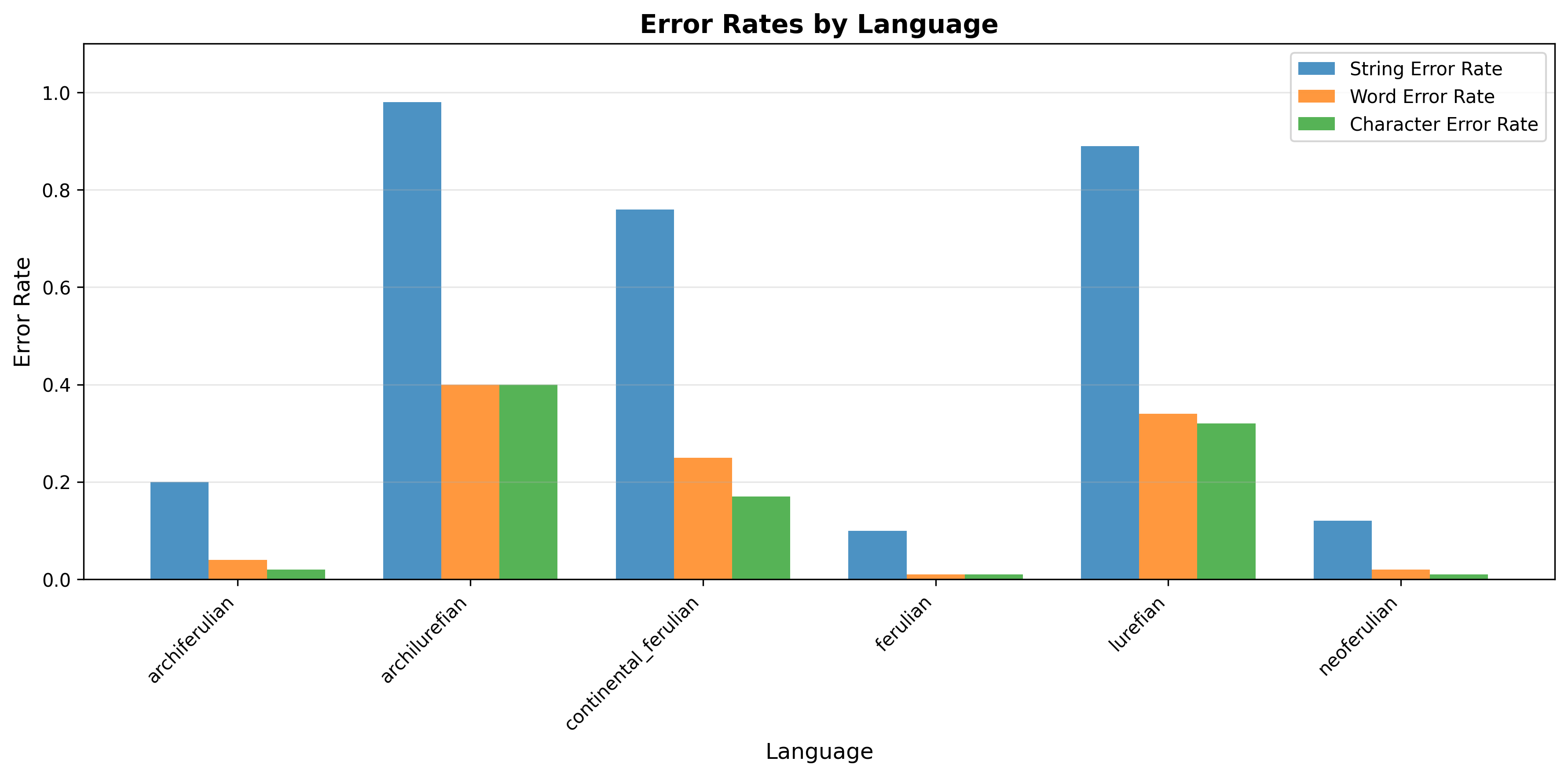} \\
\textbf{b} \\
\includegraphics[width=0.6\textwidth]{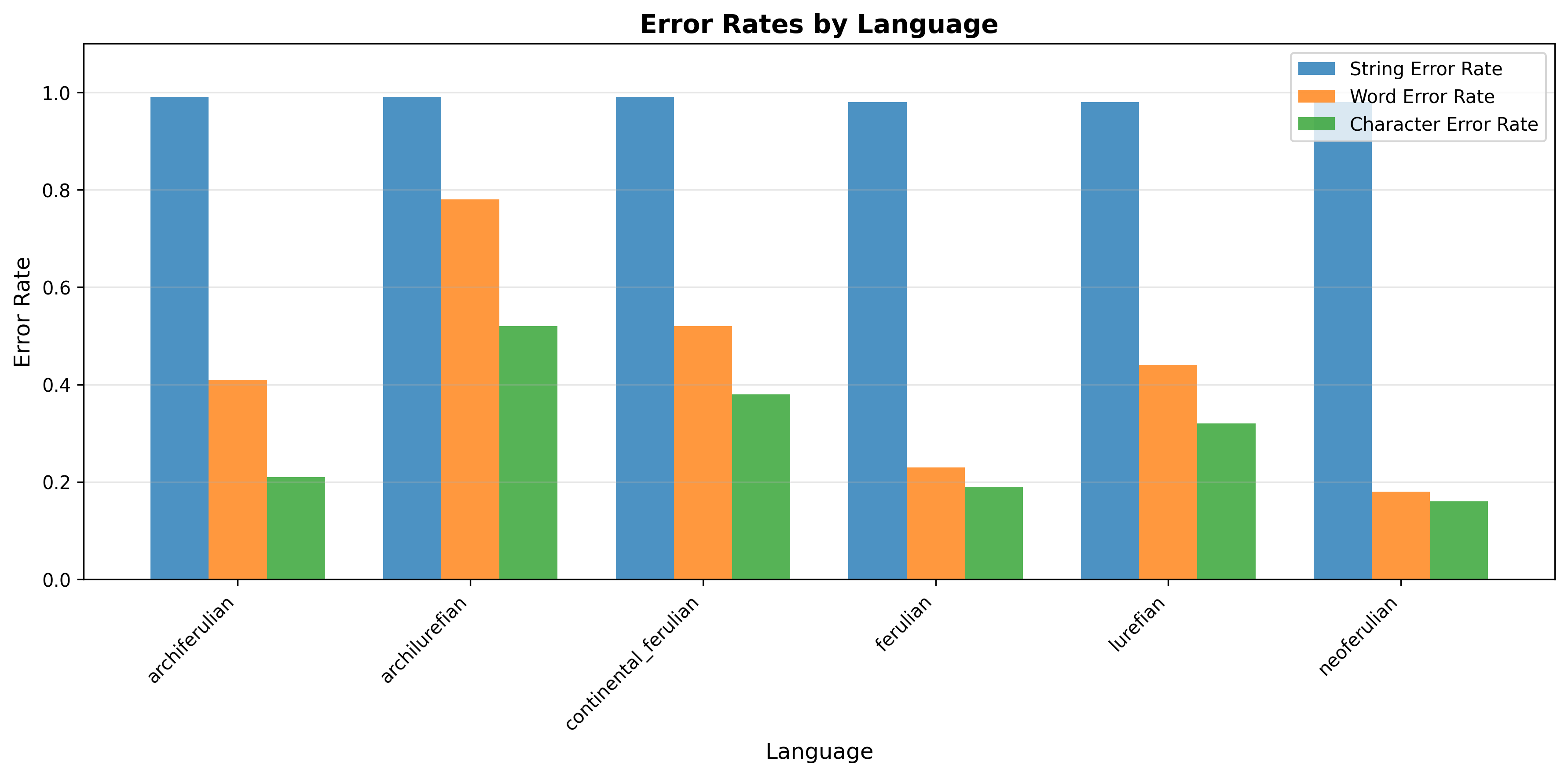} \\
\textbf{c} \\
\includegraphics[width=0.6\textwidth]{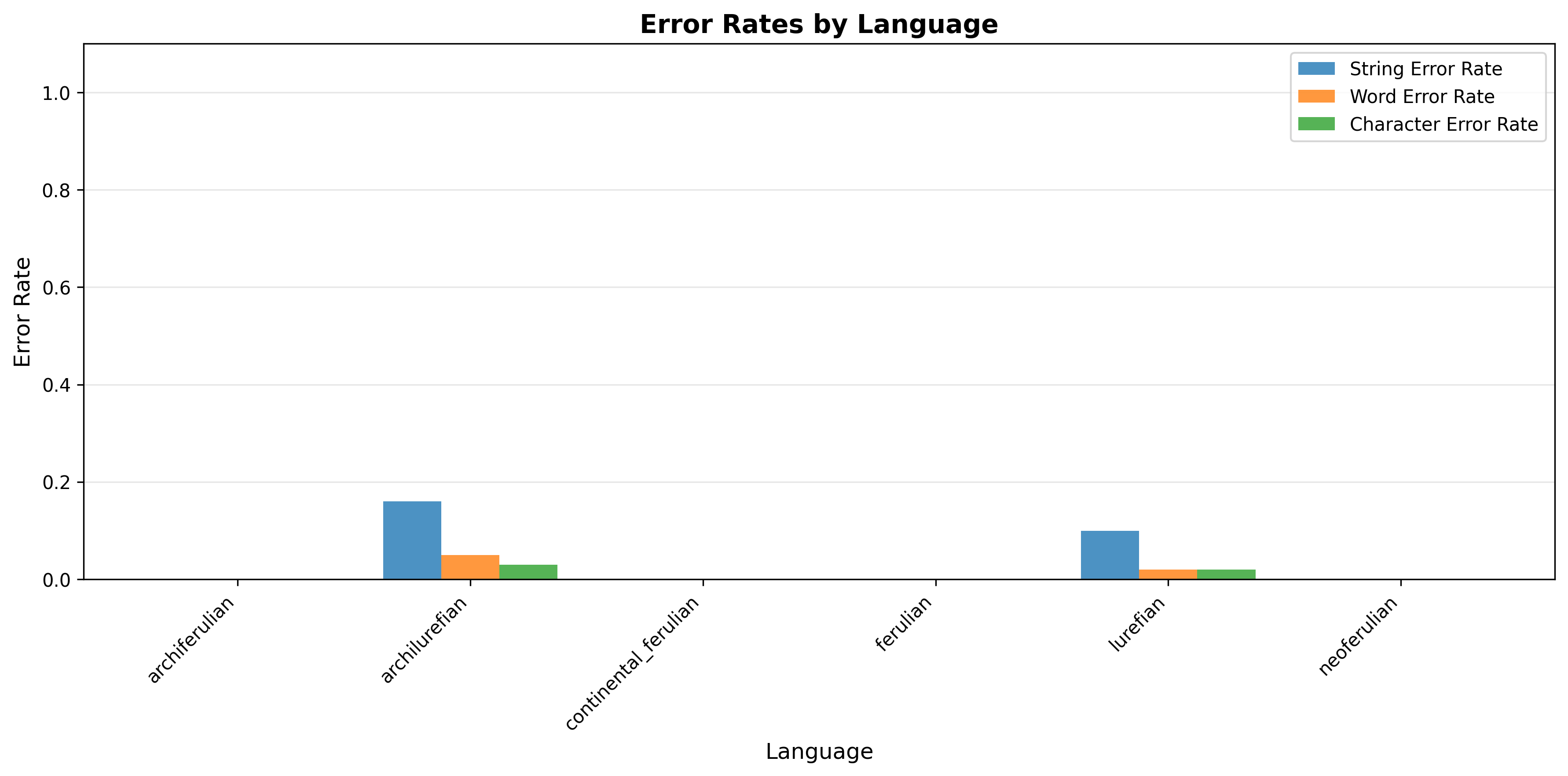} \\
\textbf{d} \\
\includegraphics[width=0.6\textwidth]{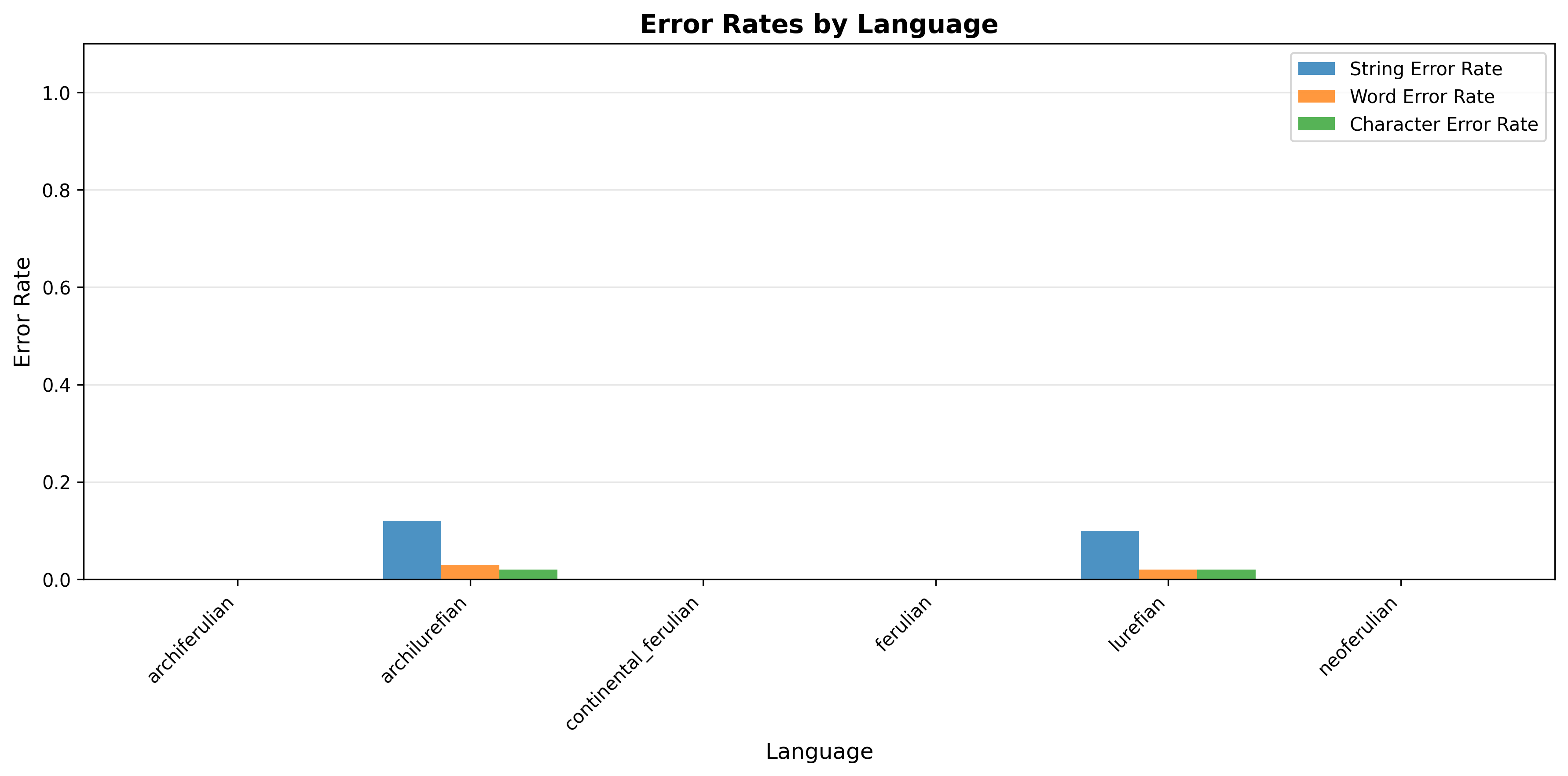} \\
\end{tabular}
\end{center}
\caption{String, word and character error rates for the six artificial number
  name systems for: (a) Claude Sonnet 4.5 (2025-09-29) reasoning; (b) Claude
  Sonnet 4.5 coding; (c) Claude Opus 4.8 reasoning; (d) Claude Opus 4.8 coding.}
\label{fig:claude}
\end{figure}

The results for Sonnet 4.5 are shown in Figure~\ref{fig:claude}a--b. They are,
as can be seen, pretty poor. For the coding variant, almost none of the numbers
are read completely correctly (b), and for the reasoning variant, three of the
six languages show very poor performance.

Miraculously, Opus 4.8 fixes practically all of this, achieving 0\% errors on
four out of five languages, and missing examples only on Lurefian and
Archilurefian, the two languages with the Igbo-like multiplier-multiplicand
reversal.  In fact, when we look at the `errors' Opus makes, in both cases those
errors point to a bug in our setup.
First, the rule to achieve the Igbo-like
reordering evidently had a bug, since rather than just \emph{one} being ordered
\emph{before} multiplicands like \emph{thousand}, and all others after, it
actually generates forms where any multiplier \emph{beginning with the digit 1},
premodifies the number: In other words, every multiplier that happens to be
spelled with an initial \emph{1}---\emph{163}, \emph{12}, \emph{150}, and so on---also triggered the exception.
For example, for the input \emph{163,080}, the
hand-built grammar produces the output labeled as `Ref', and Opus predicted the
output in `Hyp':
\begin{quote}
\ttfamily
\begin{tabular}{@{}l@{~~}cccccc@{}}
Num: & \multicolumn{6}{@{}l@{}}{163080 \g{= MPL} \p{163} \g{x} \g{MPC} \p{1000} \g{+} \g{RMD} \p{80}} \\
Ref: & [eti & molot & chelfti & vido] & kero & telfti \\
 & \multicolumn{4}{@{}c@{}}{\g{MPL}} & \g{MPC} & \g{RMD} \\
     & \p{1} & \p{100} & \p{60} & \p{3} & \p{1000} & \p{80} \\

Hyp: & {kero} & [eti & molot & chelfti & vido] & telfti \\
 & \g{MPC} & \multicolumn{4}{@{}c@{}}{\g{MPL}} & \g{RMD}  \\
     & \p{1000} & \p{1} & \p{100} & \p{60} & \p{3} & \p{80} \\

\end{tabular}
\end{quote}
Here \texttt{MPL}, \texttt{MPC}, and \texttt{RMD} are abbreviations of `multiplier', `multiplicand', and `remainder', respectively.
To compound the problem, a second bug was that for just these two
languages, examples for 100,000, 100,001 and 200,000 were accidentally omitted
from the training---thus depriving the model of a couple of crucial cases that
might have guided it to what the grammar was actually producing. But we would
argue that what the hand-written grammar is producing would be rather unnatural
as a number-name system. A system where just the number \emph{one} might behave
exceptionally as a multiplier is reasonable, and indeed there are many languages
in which a \emph{one} multiplier is treated specially. But one where any
multiplier that happens to begin with the digit \emph{1} gets exceptional
treatment is, to our knowledge, unheard of. So Opus 4.8's inferences were in
fact perfectly natural. Taking this into consideration, it seems fair to assert
that Opus 4.8 performed the task perfectly, for all languages.

Which is pretty amazing.

But at the same time, one has to ask: while the Claude family finally seems to
have achieved the stage of `grokking' the number name problem, how does this
compare to the human ability to solve the same task? In a refrain that should be
familiar by now, humans get by with far fewer resources. For many children,
learning to count and learning the number name system of one's own language
involves exposure to just one language, with no data from other languages that
could be used to infer the `general structure' of the problem. In contrast,
Claude has been exposed, we assume, to trillions of tokens of text in multiple
languages, including some texts (such as grammatical resources like Omniglot)
that discuss exactly the problem of number name formation. Beyond this, Opus 4.8
is one of the latest frontier models, incorporating (presumably) trillions of
parameters, and consuming large amounts of energy, with concomitant carbon
emission, during its training.

The problem, then, is solved---at least within the bounds of what the above
experiments test. Yet a problem still remains: how can we approach purely neural
solutions to limited-domain but constrained problems like this, without such
massive resources?  We hope that research on alternative approaches to machine
learning that go beyond the transformer-based models that are central to current
LLMs will eventually answer that question.

\starttwocolumn

\bibliographystyle{compling}
\bibliography{refs}

\end{document}